\documentclass[preprint,12pt,authoryear]{elsarticle}

\usepackage{multirow}
\usepackage{tabularx}
\usepackage{booktabs}
\usepackage{graphicx}
\usepackage{amsmath}
\usepackage{algorithm}
\usepackage{amssymb}
\usepackage{booktabs}
\usepackage{ulem}
\usepackage{array}
\usepackage{adjustbox}
\usepackage{algorithm}
\usepackage[noend]{algpseudocode}
\algrenewcommand\textproc{}
\algnewcommand{\algorithmicor}{\textbf{ or }}
\algnewcommand{\OR}{\algorithmicor}
\usepackage{mathtools}
\usepackage{hyperref}
\usepackage{xcolor}

\journal{Journal of Biomedical Informatics}

\begin{document}

\begin{frontmatter}



\title{Retrieval Augmentation of Large Language Models for Lay Language Generation}


\author{    
        Yue Guo \textsuperscript{\rm 1}\textsuperscript{*},
        Wei Qiu \textsuperscript{\rm 2}\textsuperscript{*}, Gondy Leroy\textsuperscript{\rm 3},
        Sheng Wang \textsuperscript{\rm 2}, Trevor Cohen\textsuperscript{\rm 1} \\
    \textsuperscript{\rm 1} Biomedical and Health Informatics, University of Washington  \\
    \textsuperscript{\rm 2} Paul G. Allen School of Computer Science, University of Washington\\
    \textsuperscript{\rm 3} Management Information Systems, University of Arizona\\
    \textsuperscript{*} These authors contributed equally to this work.
}


\begin{abstract}
    The complex linguistic structures and specialized terminology of expert-authored content limit the accessibility of biomedical literature to the general public. Automated methods have the potential to render this literature more interpretable to readers with different educational backgrounds. Prior work has framed such lay language generation as a summarization or simplification task. However, adapting biomedical text for the lay public includes the additional and distinct task of \textit{background explanation}: adding external content in the form of definitions, motivation, or examples to enhance comprehensibility. This task is especially challenging because the source document may not include the required background knowledge. Furthermore, background explanation capabilities have yet to be formally evaluated, and little is known about how best to enhance them. To address this problem, we introduce Retrieval-Augmented Lay Language (RALL) generation, which intuitively fits the need for external knowledge beyond that in expert-authored source documents. In addition, we introduce CELLS, the largest (63k pairs) and broadest-ranging (12 journals) parallel corpus for lay language generation. To evaluate RALL, we augmented state-of-the-art text generation models with information retrieval of either term definitions from the UMLS and Wikipedia, or embeddings of explanations from Wikipedia documents. Of these, embedding-based RALL models improved summary quality and simplicity while maintaining factual correctness, suggesting that Wikipedia is a helpful source for background explanation in this context. We also evaluated the ability of both open-soured Large Language Model (Llama 2) and closed-sourced Large Language Model (GPT-4) in background explanation, with and without retrieval augmentation. Results indicate that these LLMs can generate simplified content, but that the summary quality is not ideal.
    Taken together, this work presents the first comprehensive study of background explanation for lay language generation, paving the path for disseminating scientific knowledge to a broader audience. 
    Our code and data are publicly available at: \sloppy
    \href{https://github.com/LinguisticAnomalies/pls_retrieval}{https://github.com/LinguisticAnomalies/pls\_retrieval}.

\end{abstract}

\begin{graphicalabstract}
\includegraphics[trim={20cm 10cm 22cm 10cm},clip,scale=0.53]{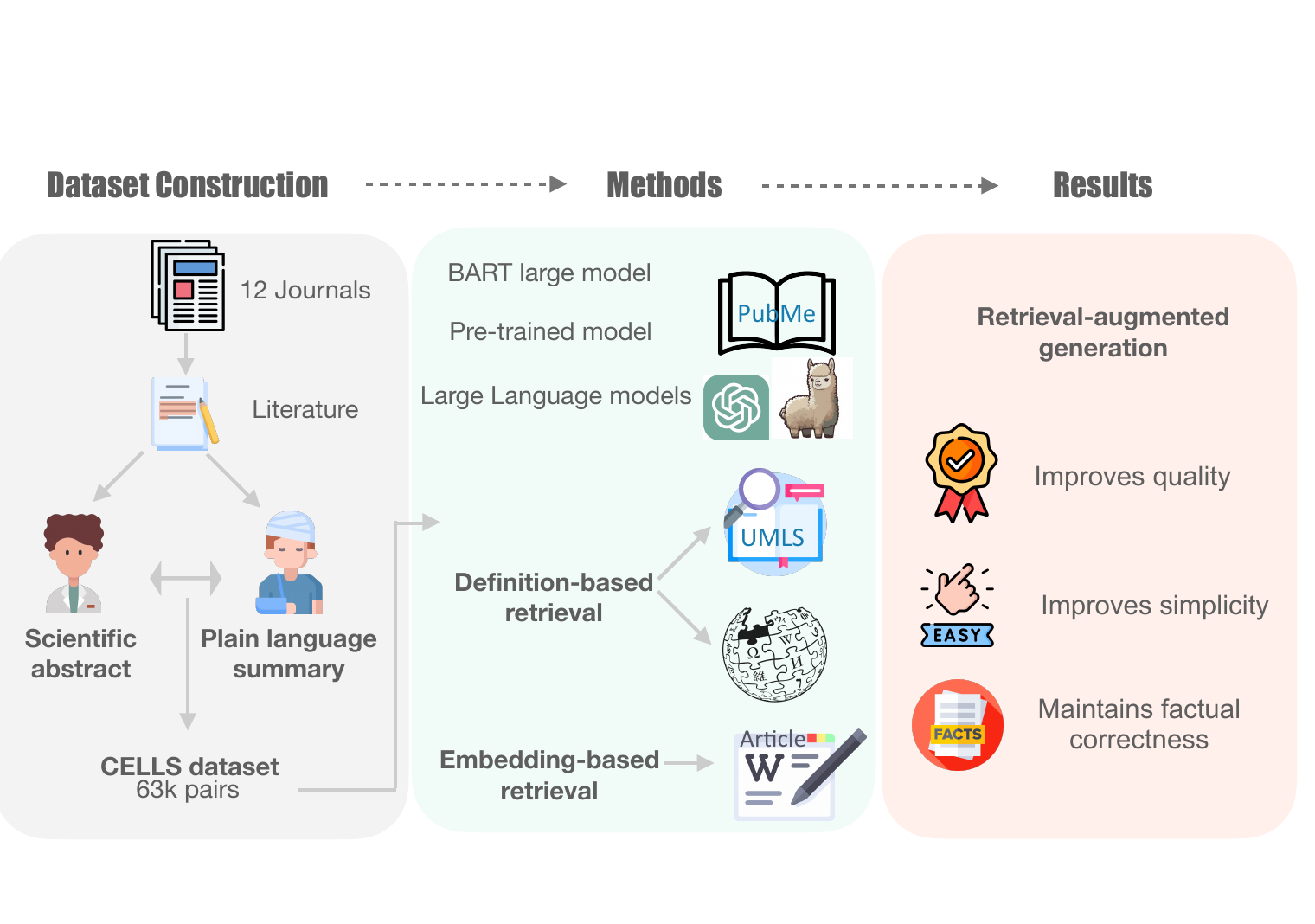}
\end{graphicalabstract}

\begin{highlights}
\item \textbf{Problem:} Automated lay summary generation can improve the accessibility of health information, but is challenging because of the need to provide background information absent in source documents.
\item \textbf{What is already known:} Current models face constraints due to corpus size, topic diversity, and untested utility of external information retrieval.
\item \textbf{What this paper adds:} We approach lay language generation by simplifying content and also generating background explanations, achieved through innovative Retrieval-Augmented Lay Language (RALL) methods. This paper also introduces CELLS, the largest (63k pairs) and most diverse (12 journals) parallel corpus for lay language generation, with a specialized subset to advance background explanation capabilities. 

\end{highlights}


\end{frontmatter}


\section{Introduction} \label{introduction}
The COVID-19 pandemic underscored the difficulties the general public faces when attempting to use scientific information to guide their health-related decisions \cite{soroya2021information, bin2021covid}. Though widely \textit{available} in scientific papers and preprints, the information required to guide health-related decision making is often not \textit{accessible}: medical jargon \cite{korsch1968gaps}, scientific writing styles \cite {kurtzman2016effective}, and insufficient scientific background \cite{crossley2014s} make this information opaque to non-experts. Consequently, there is a pressing need to deliver scientific knowledge in lay language, which has motivated research on automated generation of lay language summaries. 

Prior work has framed lay language generation as a summarization or simplification task \cite{guo2021automated, devaraj2021paragraph}. However, adapting biomedical text for the lay public includes the distinct task of \textit{background explanation}: adding external content in the form of definitions, history, or examples to enhance comprehensibility. Cognitive studies of text comprehension suggest that providing missing background information effectively improves reader comprehension, especially when readers lack the prerequisite domain knowledge to fill this in themselves \cite{mcnamara1996good}. 

Text simplification, which modifies content to improve readability while retaining its key points, has been widely studied \cite{jonnalagadda-etal-2009-towards, qenam2017text}. However, generating background information is especially challenging, because the source document may not include the required background knowledge. Furthermore, background explanation capabilities have yet to be formally evaluated, and little is known about how best to enhance them. Retrieval augmentation methods, which use information retrieval to identify additional content to inform text generation, present an intuitive fit for the need to acquire external knowledge. In the current work, we explore methods for Retrieval-Augmented Lay Language (RALL) generation, augmenting state-of-the-art text generation models with information retrieval of either term definitions from the UMLS \cite{bodenreider2004unified} and Wikipedia, or embeddings of explanations from Wikipedia documents \cite{lewis2020retrieval}. Our findings indicate that RALL models improve summary quality and simplicity while maintaining factual correctness, suggesting that general knowledge from Wikipedia in particular is a good source for background explanation. With Large Language Models (LLMs) becoming increasingly accessible, we also tested the ability of  two LLMs for background explanation: we prompted both open-source Llama 2 \cite{touvron2023llama} and closed-source GPT-4 \cite{openai2023gpt4} with and without external knowledge from Wikipedia. Results indicate that these LLMs improve simplicity but do not preserve the summary quality.

Abstractive summarization methods require source/summary pairs, with the summaries written in plain language. The limited size and topical breadth of publicly-available paired corpora constrain the scope of applicability of models trained for this task and limit the generalizability of published evaluations. Therefore, a further contribution of this work is the Corpus for Enhancement of Lay Language Synthesis (CELLS): 62,886 pairs of scientific abstracts with corresponding lay language summaries (Table \ref{example}). Summaries are written by abstract authors or other domain experts, assuring the quality of our dataset. 
CELLS is larger and more diverse than prior datasets \cite{guo2021automated, devaraj2021paragraph}, aggregating papers from 12 journals (Table \ref{journal_count}) spanning various biomedical domains. From CELLS, we derived a set of specialized paired corpora: 233,916 algorithm-aligned sentence pairs for \textit{simplification} and 47,157 scientific/lay-language pairs emphasizing novel content that is absent from scientific abstracts for \textit{background explanation} to support our research on background augmentation.

\begin{table}[htbp]
\centering
\small
\begin{tabular}{p{10cm}}
\toprule
\textbf{Abstract:} Clinical reports of Zika Virus (ZIKV) RNA detection in breast milk have been described, but evidence conflicts as to whether this RNA represents infectious virus...\\
\textbf{Summary:} \textit{Only 4 years have passed since the Zika virus outbreak in Brazil, and much remains to be understood about the transmission and health consequences of Zika infection.} \uline{To date, some case reports have detected Zika virus RNA in the breast milk of infected mothers, but the presence of a virus’ RNA does not mean that intact virus is present}...\\
\bottomrule
\end{tabular}
\caption{Example abstract/summary pair from CELLS. The lay language summary is written by the abstract's authors. There are two challenges in lay language summary generation:  generating background explanations (\textit{italicized}) and simplifying the original abstract (\underline{underlined}).}
\label{example}
\end{table}

\section{Related Work}
\subsection{Lay language summary generation}
Text summarization and simplification are two important aspects of lay language generation. Text summarization is a widely-studied research topic \cite{cohan2018discourse, cachola2020tldr, devaraj2022evaluating}. It has been a focus of research attention in the biomedical domain \cite{mishra2014text, bui2016extractive, givchi2022graph}, with applications including summarization of radiology reports \cite{cai2021chestxraybert, zhang2018learning, zhang2020optimizing}, biomedical literature \cite{wangsystematic, plaza2014comparing, cai2022covidsum} and medical dialogue \cite{chintagunta2021medically, joshi2020dr}. Several of the biomedical text simplification datasets and methods have also been reported in the literature \cite{jonnalagadda2009towards, li2020pharmmt, cao2020expertise, lu2023napss}. However, these were designed for sentence-level text simplification, rather than translation of paragraphs and longer documents into interpretable lay language. Compared to other paragraph-level lay language generation efforts \cite{guo2021automated, devaraj2021paragraph}, the current work is the first to focus on background explanation generation.

\subsection{Lay language summarization datasets}
Previous endeavors towards developing datasets for automated conversion of scientific text into lay language have been limited in scale and scope. The CL-SciSumm 2020 shared task series \cite{chandrasekaran2020overview} provided a training dataset encompassing 572 articles and corresponding author-constructed lay summaries, collated from a diverse array of scientific journals published by Elsevier. Guo et al. \cite{guo2021automated} and Devaraj et al. \cite{devaraj2021paragraph} introduce datasets of $\sim$5k scientific abstract and lay language summary pairs drawn from systematic reviews in the Cochrane Library. Goldsack et al. \cite{goldsack2022making} present $\sim$30k biomedical literature abstract pairs from PLOS and eLife. Luo et al. \cite{luo2022readability} developed a dataset from $\sim$28k biomedical abstract pairs from PLOS. Attal et al. \cite{attal2023dataset} describe the PLABA dataset, encompassing 750 pairs of abstracts, each set featuring a sentence-aligned adaptation generated by human authors.  The dataset developed for our study differs from these prior efforts in that: 1) CELLS is a large ($\sim$63K) abstract-level dataset which includes different article types besides systematic reviews; and 2) we address the need for background explanation in lay language generation, deriving a specialized subset emphasizing content that is absent from the abstracts. 

\subsection{Lay language summary generation methodologies}
Present text summarization methodologies predominantly fall into two categories: extractive and abstractive \cite{Andr2007ASO}. Extractive summarization involves ranking and selecting critical elements of the original text and combining them to form a condensed version \cite{Erkan2004LexRankGC, cheng2016neural}. In contrast, abstractive summarization introduces novel words and phrases absent from the original text \cite{gupta2019abstractive}. The necessity to provide pertinent background, explain terminology, and apply straightforward sentence structures makes lay language summarization intrinsically an abstractive task \cite{guo2021automated}. The emergence of Transformer-based approaches such as BART, T5, and PEGASUS has significantly advanced this field \cite{zhang2021leveraging, yadav2022automatic}. BART, especially when pre-trained on domain-specific data, has demonstrated strong performance in the simplification of biomedical review articles \cite{guo2021automated, goldsack2022making} and the summarization of randomized controlled trials \cite{wallace2021generating}. We employ BART as the benchmark model in the current work, including a variant with additional PubMed-specific pre-training. In addition, newer work has indicated that auto-regressive LLMs can outperform other Transformer models in lay language generation tasks \cite{goldsack2023overview}. Therefore, we also evaluated the performance of two such LLMs, including Llama 2 \cite{touvron2023llama} and GPT-4 \cite{openai2023gpt4}, on our dataset.

\subsection{Retrieval-augmented text generation}
Background explanation helps laypeople understand biomedical concepts \cite{srikanth2020elaborative}. Furthermore, experiments in cognitive psychology have shown that providing explanatory content improves the recall of readers with limited domain knowledge \cite{britton1991using, mcnamara1996good}. Information retrieval methods present an intuitive approach to identify content to inform background explanations, with established utility for clinical question answering \cite{simpson2014overview, roberts2015overview, luo2022improving}, biomedical text summarization \cite{alambo2022entity, mishra2014text, plaza2014comparing} and clinical outcome prediction \cite{naik2021literature}. There are two main categories of information retrieval methods that have been used to augment the generation of natural language text. \textit{Definition-based} retrieval methods identify terms that exist in predefined lexicons, and use their definitions to inform text generation \cite{alambo2022entity, moradi2018different}. \textit{Embedding-based} retrieval methods retrieve documents with similar low-dimensional representations, instead of depending upon lexical overlap between terms  \cite{deerwester1990indexing, cao2018encoding, guu2020retrieval, karpukhin2020dense, lewis2020retrieval}. Retrieval augmentation has been shown to improve the performance of question answering systems \cite{lewis2020retrieval}, and reduce the frequency of so-called ``hallucinations'' (statements without grounding in training data) in text generated by language models \cite{shuster-etal-2021-retrieval-augmentation}. However, these approaches have not been explored for lay language generation, despite their intuitive fit to the subtask of background explanation in particular. In the current work, we explore both definition- and embedding-based retrieval approaches and evaluate the utility of external information from the UMLS and Wikipedia for this important subtask.

\section{Materials and methods}
\subsection{The CELLS Dataset}

\begin{table}[]
\centering
\small
\begin{tabular}{@{} l@{\hspace{0.0001pt}} r@{\hspace{4pt}} c@{\hspace{4pt}}c m{.001pt} @{}}
\toprule
&& 
\multicolumn{2}{c}{\bf Length} \\

\cmidrule{3-4}
{\bf Journal} & {\bf Num.} 
& {\bf Src} & {\bf Tgt} \\     
\midrule
PNAS & 25,647 
& 227 & 124 \\
PLOS Genetics  & 8,030                    
& 256 & 192 \\    
PLOS Pathogens                                  & 7,345                    
& 260             & 193  \\
PLOS Neglected Tropical Diseases                & 7,185                    
& 315             & 198   \\  
PLOS Computational Biology                      & 7,072                    
& 253             & 188    \\ 
Cochrane                                        & 5,377                    
& 624             & 334     \\   
PLOS Biology                                    & 2,149                    
& 243             & 212    \\
Health Technology Assessment               & 557                      
& 645             & 318   \\ 
Health Services and Delivery Response      & 510                     
& 623             & 316     \\   
Public Health Research                     & 93                       
& 624             & 331     \\        
Programme Grants for Applied Research      & 78                       
& 722             & 311   \\  
Efficacy and Mechanism Evaluation          & 70                       
& 659             & 341  \\           
\bottomrule
\end{tabular}
\caption{Journals included in CELLS. The average length (token level) of lay language summaries (Tgt) is shorter than that of scientific abstracts (Src).}
\label{journal_count}
\end{table}

We present CELLS, the largest dataset of parallel scientific abstracts and expert-authored lay language summaries (LLSs) developed to date (Section \ref{subsec_cells}), offering unique opportunities to study the performance of lay language generation models. To facilitate research on key LLS generation subtasks, we have also derived subsets for simplification and background explanation (Section \ref{dataset_application}).

\subsubsection{Data compilation} \label{subsec_cells}
To develop CELLS, we manually reviewed biomedical journals and identified 19 with a LLS section (see Appendix Table \ref{journal_pls}). We collected scientific abstracts (\textit{source}) and their aligned LLSs ({\textit{target}}) from these journals. We excluded abstracts where LLSs are not associated with a full-length paper (i.e., LLS in a separate section for the journal's website or social media feed) that required extensive human inspection. After further excluding non-biomedical topics, we obtained 75,205 pairs of abstracts and LLSs. To ensure data quality, we identified outliers using source-target lexical similarity and length. As a result, we excluded pairs from eLife, Annals of the Rheumatic Diseases, and Reproductive Health. This left a set of 62,886 source-target pairs from 12 journals. 

\subsubsection{Dataset applications} \label{dataset_application}
Using CELLS, we developed three evaluation tasks:
\paragraph{Lay language generation}
For this task, we used the full-length scientific abstract and LLS pairs in CELLS for abstract-level lay language generation. As mentioned in Section \ref{introduction}, this task requires paragraph-level simplification, summarization, and background explanation to produce understandable summaries for laypeople. The following tasks focus on two of these challenges: abstract simplification and background explanation generation.

\begin{figure*}[htp]
    \centering
    \begin{adjustbox}{center}
    \includegraphics[scale=1]{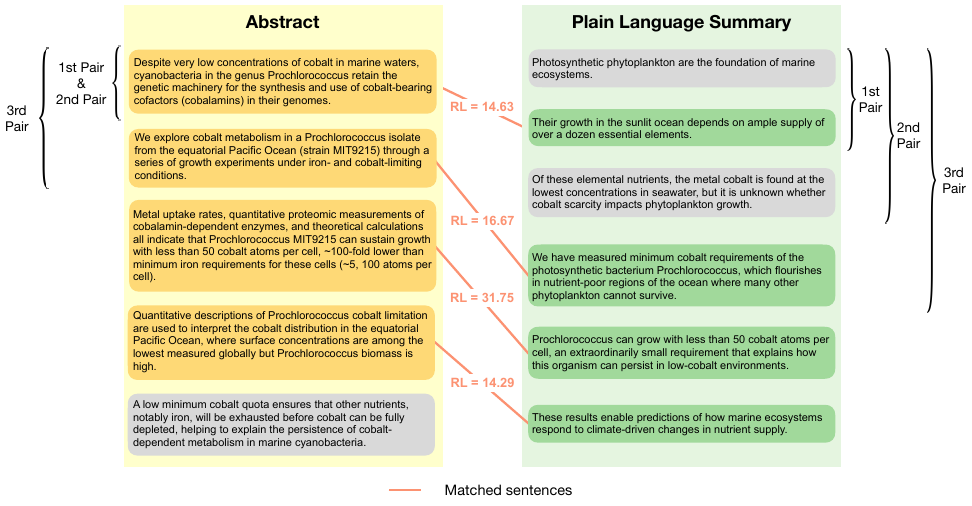}
    \end{adjustbox}
    \caption{An example application of the GPSS algorithm. RL indicates the F1 score from ROUGE-L between the sentences in the abstract and plain language summary. For the background explanation subset, we combined unaligned target sentences (grey blocks) with proximal aligned sentences (green blocks). The example presented illustrates the generation of three paired examples (``pair'') for the background explanation subset. All three pairs include the initial explanatory content that precedes the first matched sentence (RL = 14.63), as well as the sentence in the lay language summary that matches it. The second pair also includes the explanatory content after this matched sentences, and the third pair adds the following matched sentence also (i.e. the second sentence in the source abstract, and the lay language summary sentence that aligns with it). These combinations allow for the possibility that added content may relate to the preceding, or the subsequent sentence.}
    \label{fig:GPSS}
\end{figure*}

\paragraph{Simplification} 
Paragraph-level simplification fits the lay language generation task, but simplification is difficult to isolate because of the frequent insertion of background explanations. To focus on sentence-level simplification as a subtask, we developed a Greedy Paired Sentence Search (GPSS) algorithm (Algorithm \ref{GPSS_pseudocode}) to align sentences from the abstracts and LLSs. The underlying idea is to identify matched source and target sentences based on lexical overlap and sentence sequence. An example is provided in Figure \ref{fig:GPSS}.  
After applying GPSS, each source and target sentence was labeled as ``matched'' or ``unmatched'', resulting in a large set of 233,916 matching  abstract- and LLS-derived sentence pairs for simplification.

\begin{algorithm}[hbt!]
\caption{Greedy paired sentences search (GPSS) algorithm}\label{GPSS_pseudocode}
\hspace*{\algorithmicindent} \textbf{Input: }SRC -- the list of sentences in the source abstract, TGT -- the list of sentences in the target abstract
\hspace*{\algorithmicindent} \textbf{Output: } P -- the set including the indices of the paired source and target sentences
\begin{algorithmic}[1]
\Function{GPSS(src\_start, src\_end, tgt\_start, tgt\_end, score)}{}
\If {src\_start $>$ src\_end \OR tgt\_start $>$ tgt\_end}
\State \Return $\emptyset$ 
\EndIf
\State src\_max,tgt\_max $\gets$ argmax$_{i,j}$(score[i,j], src\_start $\leq$ i $<$ src\_end, tgt\_start $\leq$ j $<$ tgt\_end)
\State pairs $\gets$ \{(src\_max, tgt\_max)\}
\State pairs $\gets$ pairs $\cup$ GPSS(src\_start, src\_max-1, tgt\_start, tgt\_max-1, score)
\State pairs $\gets$ pairs $\cup$ GPSS(src\_max+1, src\_end, tgt\_max+1, tgt\_end, score)
\State \Return pairs
\EndFunction
\State N$_{src}$ $\gets$ the number of sentences in SRC
\State N$_{tgt}$ $\gets$ the number of sentences in TGT
\For{i $\gets$ 0 to N$_{src}$-1}
\For{j $\gets$ 0 to N$_{tgt}$-1}
\State S[i,j] $\gets$ ROUGE-L(TGT[j], SRC[i])
\EndFor
\EndFor
\State P $\gets$ GPSS(0, N$_{src}$, 0, N$_{tgt}$, S)
\end{algorithmic}
\end{algorithm}

\paragraph{Background explanation} As mentioned in Section \ref{introduction}, adding explanations is a common strategy to enhance comprehension in `Background' section. To support this research, we derived a large-scale paragraph-level dataset that emphasizes the insertion of novel content, i.e., background explanations. Focusing on explanation requires a reliable approach to extract sentences containing additional content in the LLS `Background' section. Human annotation is reliable but costly, and exhaustive annotation of the 62,886 pairs in CELLS is infeasible. Therefore, we obtained paired source/target sub-paragraphs with the aforementioned GPSS algorithm. 
After applying GPSS, we considered the unaligned (``unmatched'') sentences as putative \textit{explanations}. We targeted the Background section, but section headers are unavailable for most abstracts. We therefore conducted human annotation to examine the utility of different empirically-defined boundaries, and the presence of external information. Fifty randomly selected abstracts from CELLS were annotated by two annotators: one medical student and one graduate student without medical training but with good familiarity with the dataset. Cohen's Kappa among the  annotators is 0.74, indicating substantial agreement \cite{artstein2008inter}. Informed by the results of the annotation process, we selected the 2nd pair (refer to Figure~\ref{fig:GPSS}) to demarcate the `background' section, given its superior integration of background and external information. Further details can be found in \ref{background_annotation}. Overall, we extracted 47,157 source/target pairs for background explanation \footnote{The term "background explanation" refers to specific sentences found within a Background section, but not every sentence in this section qualifies as a "background explanation." Instead, a background explanation is specifically defined as unmatched sentences serving explanatory purposes.}. 

\begin{figure*}[h!]
    \centering
    \begin{adjustbox}{center}
    \includegraphics[trim={0 6pt 0 10pt},scale=0.8]{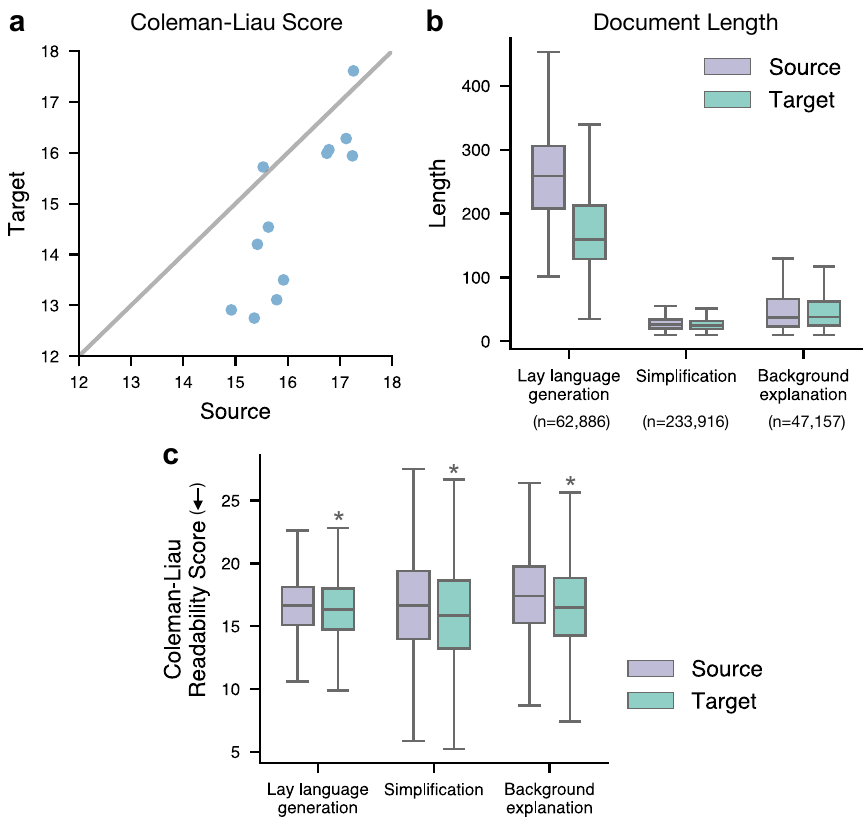}
    \end{adjustbox}
    \caption{Dataset analysis. a, source and target Coleman-Liau readability scores for the 12 journals included in CELLS. Each dot represents one journal. Lower score indicates text is easier to read. b,c, Average length and Coleman-Liau readability score for source and target text for three tasks (i.e., lay language generation, simplification and  background explanation). On average, target text is shorter and easier to read
    for all three tasks. ``*'' indicates that the score of the target significantly lower than that of the source with p-value $<$ 0.05 (paired t-test).}
    \label{fig:task_source_target_score}
    \vspace{-.2cm}
\end{figure*}

\subsection{Human Validation of Dataset}
To ensure the robustness of the dataset used for background explanation and simplification, two expert annotators (same as above)  assessed 250 paragraph pairs from the background explanation subset and 500 pairs from the simplification subset. The annotators were tasked with evaluating the pairs from the background explanation by: 1) confirming their presence within the background section; 2) identifying any external data not originally in the source; 3) classifying any external information as either definition, motivation, or example (detailed definitions of the categories can be found in Section~\ref{sec:examples}); and 4) discerning whether the target and source information are aligned, where alignment is defined by the presence of common entities, i.e., at least one shared "triple" (A triple consists of three components: A subject, a predicate, and an object). Our annotators determined that 92.8\% of pairs were situated in the background section and 62.8\% included external information. Among the identified external information, 76.4\% were motivations, 38.2\% were definitions, and 10.8\% were examples (these labels are not mutually exclusive). In addition, 87.2\% of background explanation pairs were found to be in alignment. For the simplification subset, we focused on the alignment of paired sentences and found a 68.6\% alignment. The challenge of sentence-level alignment in scientific summaries remains an active area of investigation \cite{krishna2023longeval}, emphasizing the ongoing need for the advancement of alignment algorithms. Taking into account the inherent complexities of sentence alignment and external information detection, we consider the observed alignment and external rates to be within acceptable bounds for the purpose of the current work. However, these results also reinforce the need for further research on sentence alignment.

\subsubsection{Dataset analysis} \label{dataset_analysis}
Dataset statistics are shown in Table \ref{journal_count}. CELLS covers various topics including genetics, pathogens, neglected tropical diseases, computational biology, health services, and biomedical research. This diversity of topics and journals provides opportunities to study model generalizability. Background explanations notwithstanding, the average length of the source (professional language) is longer than the target (plain language summary) for each journal. The readability scores for each journal are shown in Figure \ref{fig:task_source_target_score}a. The Coleman-Liau readability score indicates the estimated years of education required to understand a piece of text. Most of the average readability scores for the target are lower than those for the source, indicating that the target LLSs are generally easier to understand. 

Figures \ref{fig:task_source_target_score}b and \ref{fig:task_source_target_score}c show lexical features of CELLS components for three tasks. On average, LLSs are shorter than corresponding scientific abstracts. Although the readability of both source and target texts is at the college level \cite{karavcic2019languages}, the difference in readability between them is statistically significant (paired t-test), indicating LLSs are easier to understand than source text. 

We randomly split the dataset into 45,280; 11,295; and 6,311 abstract/LLS pairs as the train, validation, and test sets respectively.

\subsection{Methods}
We investigated the performance of language models with intermediate pre-training (i.e. further pre-training on in-domain text) and retrieval-augmented lay language generation (RALL).
\subsubsection{In-domain pre-training for simplification}
\label{sec:inter-pretraining}
Abstractive models are more applicable than extractive ones for our tasks since extractive summaries are written in the same professional language as their source documents. Therefore, we applied a state-of-the-art abstractive summarization model -- \textit{BART} \cite{lewis2020bart} -- to our tasks. BART uses hidden state representations of text sequences that are encoded bi-directionally (as is the case with BERT \cite{devlin2018bert}) to inform a decoder model that predicts the next word in the sequence (as is the case with GPT-series and related models e.g. \cite{brown2020language}). During semi-supervised pre-training of the model, input sequences are perturbed with a range of transformations (for example, some tokens may be masked), and the model attempts to reconstruct the original sequence. As such, BART has both the ability to encode hidden state representations that take an entire sequence into account, and a convenient mechanism to generate text in response to an input sequence. Once the model has been pre-trained on unlabeled text, it can be fine-tuned for particular tasks, such as summarization, by training it to generate target sequences in response to source sequences. We adopted a $\text{BART}_{\text{Large}}$ model that has been fine-tuned for general-domain text summarization on the widely-used CNN/DM summarization dataset (Cable News Network / Daily Mail \cite{nallapati2016abstractive}) as our baseline. To be concise, we use \textit{Vanilla} to denote the BART-Large-CNN model in the following text. Due to the complexity of our task and the relatively small size of our data, we employed intermediate pre-training - further semi-supervised pre-training on additional unlabeled in-domain text - to attempt to improve the performance of BART. Previous work shows that adaptive pre-training with domain-relevant unlabeled data can improve model performance \cite{gururangan2020don}. Therefore, we further pre-trained the BART model on a corpus from the biomedical domain. We first perturbed 300K PubMed abstracts\footnote{https://www.kaggle.com/cvltmao/pmc-articles} by text substitution and sentence shuffling, and trained the BART model to reconstruct the original text. The pre-trained model was then further fine-tuned for the tasks of summarization, sentence simplification and background explanation, using our datasets.

\subsubsection{Definition-based explanation retrieval}



As the source documents in our dataset may omit required background knowledge, models should be able to retrieve relevant background knowledge from external sources. We evaluated two approaches to retrieving this knowledge.


The definition-based retrieval model uses a straightforward method to add explanations of terms to the text, by identifying definitions of terms that exist in a predefined lexicon. In our experiments, we used datasets derived from the UMLS \cite{bodenreider2004unified} and Wikipedia to augment the context of the source document. The UMLS includes medical term (entity $e$) and definition ($d$) pairs $\mathcal{D}_u=\{(e_i, d_i)\}$. 
For each source document $s$, we used Scispacy \cite{neumann2019scispacy} to identify the expression of normalized UMLS terms ${e_{u_1}, e_{u_2}, ..., e_{u_m}}$ in $s$. Then we added corresponding UMLS term definitions $d_{u_1}, d_{u_2}, ..., d_{u_m}$ to $s$ to obtain $\hat{s}$. The Wikipedia dataset includes keyword ($e$) and definition ($d$) pairs $\mathcal{D}_w=\{(e_i, d_i)\}$. For each source document $s$, we applied KeyBERT\footnote{https://github.com/MaartenGr/KeyBERT}, which uses BERT embeddings and cosine similarity to find the sub-phrases in a document most similar to the document itself, to identify three keywords $(e_{w_1}, e_{w_2}, e_{w_3})=\text{KeyBERT}(s)$. 
We obtained the definitions of those keywords, $d_{w_1} d_{w_2}, d_{w_3}$, from the Wikipedia dataset and added them to the end of the source document, $s$, to obtain $\hat{s}$. Lastly, we fine-tuned the BART model using $\hat{s}$.

\subsubsection{Embedding-based explanation retrieval}
For the embedding-based retrieval method, we adopted a state-of-the-art dense retrieval model to augment the source with related documents from an external set. Specifically, we applied the retrieval-augmented generation (RAG) model \cite{lewis2020retrieval} using another Wikipedia-derived dataset $Z_w$ of 21M 100-word documents $z$. The retrieval component $p_{\phi}(z|s) \propto \rm{exp}(\textbf{d}(z)^{\rm T}\, \textbf{q}(s))$ is based on the Dense Passage Retriever (DPR) \cite{karpukhin2020dense}, where $\textbf{d}(z)=\text{BERT}_d(z)$ and $\textbf{q}(s)=\text{BERT}_q(s)$ are the representations of the documents (in our case, the source document to be summarized and the Wikipedia documents that provide candidates for retrieval) produced by two $\text{BERT}_{\text{Base}}$ encoders. The DPR model retrieves the top $k$ documents $z$ with the highest prior probability $p_{\phi}(z|s)$ using the Maximum Inner Product Search method \cite{johnson2019billion}. After applying DPR, we concatenated the source $s$ and the retrieved content $z$ as the input. We used the RAG-Sequence model whose generator produces the output sequence probabilities for each concatenated document:
$$p(t|s) \approx \sum_{\mathclap{z \in top\text{-} k(p_\phi(\cdot|s))}}p_{\phi}(z|s)p_{\theta}(t|s,z).$$
$p_{\theta}(t|s,z)$ is the generator, and we used BART for this purpose. As can be seen from the formula, both the content of the retrieved documents ($z$) and their probabilities of retrieval ($p_{\phi}(z|s)$) inform the generated text. During training, we fixed $\text{BERT}_d(\cdot)$, and only fine-tuned $\text{BERT}_q(\cdot)$ and the BART generator. 

\subsection{LLMs}
To evaluate the performance of LLMs in generating background explanations or plain language summaries, we utilized Llama 2 \cite{touvron2023llama} and GPT-4 \cite{openai2023gpt4}. We explored two prompts: 1) \texttt{"Summarize in plain language: {input}"}, and 2) \texttt{"Summarize in plain language, providing necessary explanations: {input}"}. To further assess the impact of the retrieval-augmented approach on LLMs, we established two settings for input: the source alone and the source combined with Wikipedia definitions as identified using KeyBERT.





\subsection{Experiments}
\subsubsection{Experimental setup}
All experiments except LLMs were run using a single NVIDIA Tesla V-100 GPU. Models were developed using PyTorch \cite{paszke2019pytorch}. We used the Fairseq\footnote{https://github.com/pytorch/fairseq} BART implementation, and the HuggingFace Transformers Library \cite{wolf2019huggingface} to implement the RAG model. For RAG, we retrieved the top 5 
documents for each input. The maximum length of generated texts was set to 700 for paragraph-level lay language generation and 150 for background explanation and sentence-level simplification. Other hyper-parameters were set to their default values.

We used the \texttt{Llama-2-70B-chat}\footnote{Model: https://ai.meta.com/llama/} model for Llama 2 \footnote{Implementation: The model was quantized to 4 bits using OPTQ as implemented in https://github.com/PanQiWei/AutoGPTQ, and hosted on a local server using https://github.com/turboderp/exllama for inference} and GPT-4 \footnote{Implementation: https://openai.com} for the GPT model. The generation process was configured with a maximum length of 150 tokens. All other parameters were set to their default values.

\subsubsection{Evaluation}
\paragraph{Automated evaluation}
We first evaluated generation quality using ROUGE-L \cite{lin2004rouge}, BERTScore \cite{bert-score}, BLEU \cite{papineni2002bleu}, and METEOR \cite{banerjee2005meteor}\footnote{Implementation: \cite{fabbri2020summeval} BERTScore hash code: \texttt{bert-base-uncased\_L8\_no-idf\_version = 0.3.12(hug\_trans=4.27.3)}} to compare generated text to professionally-authored plain language target text. ROUGE-L depends on \(n\)-gram overlap, while BERTScore uses the similarity between embeddings and as such may be less sensitive to differences in word choice between human-authored and automatically-generated LLS. BLEU computes \(n\)-gram precision of generated text against target texts, including a brevity penalty. METEOR employs a relaxed matching criterion based on the F-measure, and addresses the exact match restrictions and recall consideration of BLEU.\footnote{Please see BLEU and METEOR scores in Appendix.} The Coleman-Liau readability score \cite{coleman1975computer} assesses the ease with which a reader can understand a passage, and word familiarity \cite{leroy2014effect} measures the inverse document frequency of unigrams in text using frequency counts from Wikipedia. \textit{Lower} Coleman-Liau score and word familiarity indicate that text is \textit{easier} to read.\footnote{The familiarity measure is derived from \textit{inverse} document frequency which is higher for rare terms, so \textit{lower} familiarity scores indicate the use of \textit{more familiar} words.}

To directly evaluate how representative of an LLS the generated text is, we trained a RoBERTa \cite{liu2019roberta} model to classify the source of sentences from the original abstracts and LLSs. Specifically, we used the paired source-target sentences from the GPSS algorithm with a sentence length between 10 and 150 words. The input to the RoBERTa model is a sentence and the label is 0 for a source sentence (from a scientific abstract) and 1 for a target sentence (from a LLS). The RoBERTa model achieved an AUROC of 0.83 and an F1 score of 0.74 on the held-out test set, demonstrating that there are detectable and generalizable differences in language use by the intended audience. As the model is trained to output a higher prediction for a target sentence (i.e., a LLS sentence), we used the prediction of the model to evaluate how ``plain'' the input text is. We refer to the predicted probability of this model as the ``Plainness Score''. A \textit{higher} Plainness Score indicates that the text is more representative of an LLS.

\paragraph{Human evaluation}
We set up our human evaluation similarly to \cite{guo2021automated}, providing pairs of source and summary text to the human evaluators, where the summary is either expert-written or generated by one of our two best-performing BART models and two GPT-4 models (evaluators were not informed which summaries were human-authored). We asked human evaluators to rate the summary for grammatical correctness, meaning preservation, understandability, factual correctness, and the relevance of external information, each on a 1-4 point Likert scale (1-very poor, 4-very good). Questions can be found in \ref{human_evaluation_question}. The study was considered exempt upon institutional IRB review. Twelve evaluators were recruited using an institutional NLP interest group channel.  Each of them annotated four examples from the test set for background explanation. Three evaluators reviewed each example. All the evaluators have at least an undergraduate degree, lack specialized biomedical training, and half speak English at home. The Krippendorff's alpha coefficient \cite{krippendorff1970estimating} was 0.40 for the four background explanation texts among all evaluators. Krippendorff's alpha coefficient measures multiple inter-rater agreements in ordinal data, and values range from 0 to 1. Considering the subjectivity of the task and the multiple choices per question, we considered this level of agreement to be acceptable.

\section{Results}
\begin{figure*}[h!]
    \centering
    \begin{adjustbox}{center}
    \includegraphics[scale=0.9]{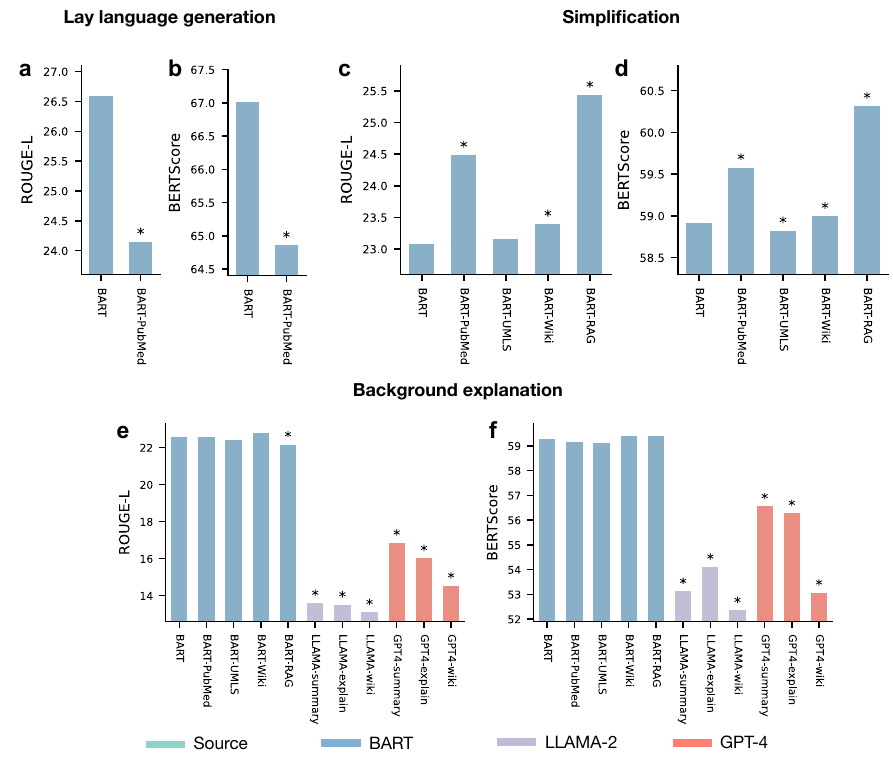}
    \end{adjustbox}
    \caption{Models' performance in text generation. We used the F1 score of ROUGE-L and BERTScore to evaluate the generation quality of models on lay language generation, simplification, and background explanation tasks. P-values obtained through the t-test are employed to evaluate the performance of various models compared to the Vanilla model (BART). * indicates statistical significance with Bonferroni-Holm correction for multiple hypothesis testing \cite{holm1979simple}. }
    \label{fig:generation_quality}
    \vspace{-.25cm}
\end{figure*}
We experimented with five models using BART: the base (\textit{Vanilla}) model, \textit{Vanilla} further pre-trained on PubMed abstracts (\textit{PubMed pre-trained}), \textit{Vanilla} with UMLS (\textit{UMLS definition-based retrieval}) and Wikipedia (\textit{Wiki definition-based retrieval}) definition-based retrieval, and \textit{Vanilla} with embedding-based retrieval using Wikipedia (\textit{Wiki embedding-based retrieval}). Additionally, we experimented with three prompts using LLMs for background explanation: "Summarize in plain language: {input}" (\textit{summary}), "Summarize in plain language, providing necessary explanations: {input} (\textit{explain}), and "Summarize in plain language: {input with Wiki definition-based retrieval} (\textit{wiki}).

\subsection{RALL improves generation performance}
We first evaluated the text generation performance of our models (Figure \ref{fig:generation_quality}), using ROUGE-L and BERTScore to compare generated LLS text to the corresponding human-authored lay language text (the target) for a given abstract or sentence (the source). Due to the input length limitation of the BERT (512 tokens) and BART (1024 tokens) models, we did not perform retrieval-augmented generation for the abstract-level lay language generation task (Figure \ref{fig:generation_quality}, 1st panel). However, for this task, pre-training on unlabeled data was not helpful.

We next compared text generation performance on the sentence-level simplification task (Figure \ref{fig:generation_quality}, 2nd panel). Results indicate that the PubMed pre-trained model achieved better performance than the Vanilla model, suggesting that pre-training on domain-specific unlabeled data is helpful for sentence-level simplification, which aligns with the results from our prior work \cite{guo2021automated}. It can also be observed that the models with information retrieval from Wikipedia (Wiki definition- and embedding-based retrieval) achieve higher ROUGE-L scores than the Vanilla model, suggesting that retrieving this external information may also be helpful for text simplification tasks. One reason for this may be that Wikipedia articles target a broader audience than the intended audience of specialized biomedical literature, and are therefore written to be more accessible.

For background explanation  (Figure \ref{fig:generation_quality}, 3rd panel), the PubMed pre-trained model only shows marginal improvements, suggesting that pre-training on domain-specific unlabeled data is helpful but insufficient for this challenging task, perhaps because the authors of biomedical literature assume expert knowledge on the part of their readers and therefore seldom include the explanatory content that a non-expert reader might require. Furthermore, the BART models with retrieval from the Wikipedia dataset (Wiki definition- and embedding-based retrieval) achieved higher BERTScore, establishing the benefits of information retrieval techniques for background explanation with BART. ROUGE-L results show a smaller advantage for Wiki definition-based RALL generation, and unlike with BERTScore this advantage is not apparent for embedding-based RALL methods. With auto-regressive LLMs ROUGE and BERTscore results are remarkably consistent: GPT-4 outperforms Llama 2 with all three prompts; however, both models are notably outperformed by BART-based architectures. Our results suggest that prompting exclusively for summarization yields superior outcomes compared to combining summarization with explanation. The weakest performance is observed when using the Wiki definition-based retrieval source combined with summarization prompting. This disparity could be attributed to our reliance on zero-shot LLMs, whereas BART benefits from fine-tuning. This suggests there may be avenues for improvement, such as exploring few-shot learning approaches within LLMs for background explanation, or using low-rank adaptation techniques to further improve auto-regressive LLM performance. To offer a comprehensive view of performance, BLEU and METEOR scores from the test set are also presented in Appendix Figure~\ref{appfig:generation_quality_bleu}. The observed BLEU patterns are consistent with those for ROUGE and BERTScore, while METEOR highlights advantages for `explain' LLM queries and BART-RAG.

In acknowledgement of the limitations of our GPSS algorithm, where the background explanation doesn't always pair paragraphs with external content and the simplification subset doesn't always produce aligned pairs, we provide results from the `gold' annotated dataset. These are available in Appendix Figure~\ref{appfig:generation_quality_validate_rouge} and Appendix Figure~\ref{appfig:generation_quality_validate_bleu}.  This `gold' background explanation subset features pairs in alignment with external content, while the `gold' simplification subset showcases aligned pairs. The results derived from these `gold' validated subsets are consistent with those from the test set, indicating that the observed patterns are not attributable to errors in algorithmic alignment.

Overall, our results suggest that both biomedical domain pre-training and information retrieval are helpful for background explanation and sentence-level simplification. Furthermore, the information retrieval models based on BART using Wikipedia produced text that was most similar to human-authored lay language. This indicates that general domain information written for a broader audience (e.g., Wikipedia) is a good resource for background explanation generation using BART.  

\subsection{RALL improves text interpretability}
\begin{figure}[h!]
    \centering
    \includegraphics[scale=1]{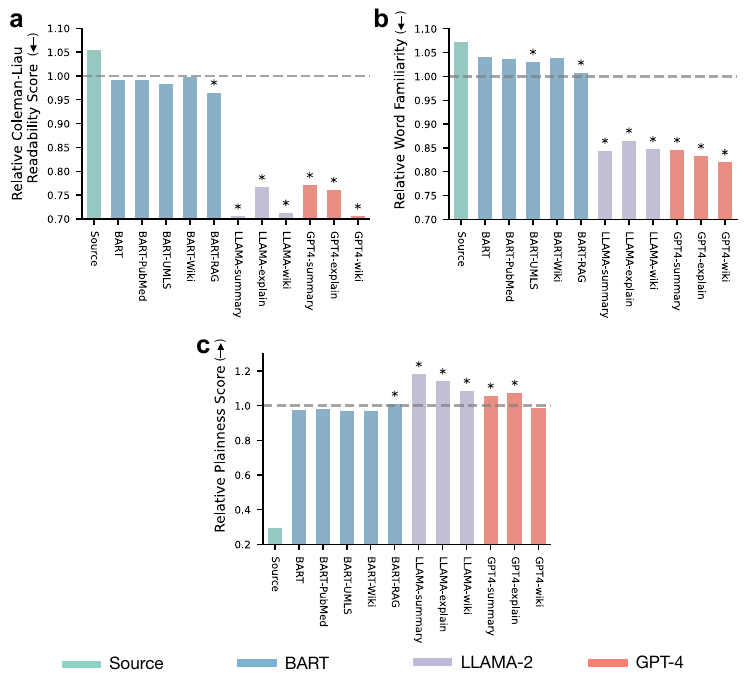}
    \caption{Readibility, familiarity and plainness of the background explanation subset, relative to professionally-authored lay language text. (a) Relative Coleman-Liau readability score, (b) word familiarity, and (c) plainness score of the source and models' generated text. The relative score is calculated by dividing by the score of the target text. A lower readability score and word familiarity indicate that the text is easier to read (values below the dashed line are lower than those from professionally-authored plain language). A higher Plainness Score indicates that the text is more representative of an LLS. P-values obtained through the t-test are employed to evaluate the performance of various models compared to the Vanilla model (BART). * indicates statistical significance with Bonferroni-Holm correction for multiple hypothesis testing \cite{holm1979simple}.}
    \label{fig:test_results}
    \vspace{-.25cm}
\end{figure}

We next evaluated the interpretability of the generated text using the data from the background explanation subset (Figure \ref{fig:test_results}). Existing text interpretability metrics consistently return better scores for the models' outputs than for the source text. Of note, the Coleman-Liau readability scores of the models' outputs are even lower than those of the target text (Figure \ref{fig:test_results}a.) This indicates that our datasets help the BART model to generate more straightforward and readily interpretable text. We also found that the retrieval-augmented BART models performed well in this interpretability evaluation, suggesting that the UMLS and Wikipedia datasets may be easier to understand than professional-language abstracts. Overall, the embedding-based RALL model, which used Wikipedia as a source, had the best readability, familiarity, and plainness scores. These results further support the utility of retrieval augmentation for lay language generation, suggesting it can benefit the style of generated text, as well as its content. For LLMs, the model outputs consistently score well across all metrics. Outputs generated with the Wiki-based definition retrieval source show improved results in the readability score and relative word familiarity compared to those without it. However, this advantage doesn't extend to the plainness score.

\subsection{Human evaluation}
\begin{figure}[h!]
    \centering
    \includegraphics[scale=0.5]{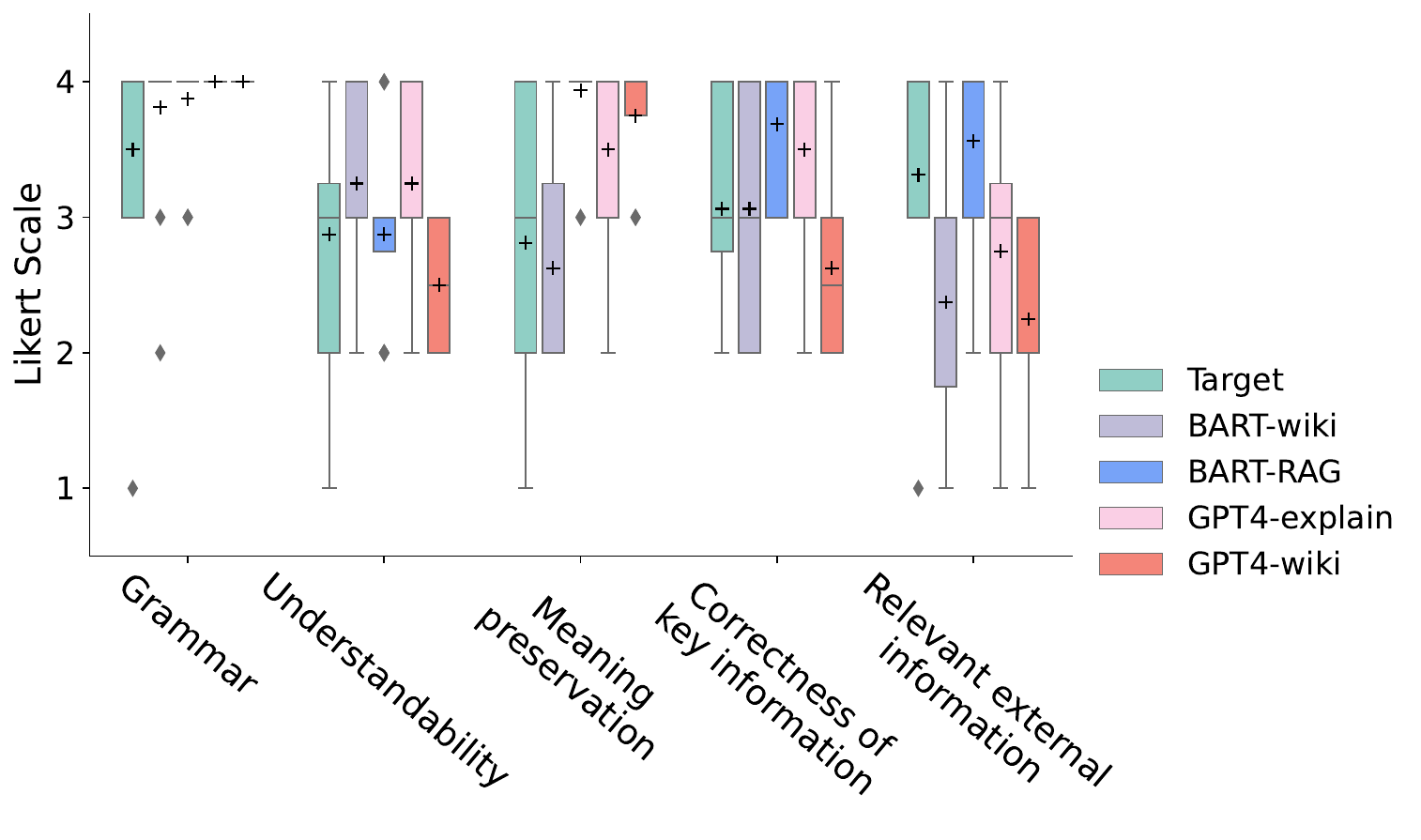}
    \caption{Human evaluation results for four generated texts from the background explanation task. Each text was assigned to four raters. For the Likert scale, 1 is very poor, and 4 is very good. ``+'' indicates the mean.}
    \vspace{-.3cm}
    \label{fig:human_evaluation}
\end{figure}
Figure \ref{fig:human_evaluation} shows the human rating scores across four pairs of target and generated texts. Evaluators generally rated generated background explanations higher than those from the expert-generated LLS. It is interesting to note that the Wiki definition-based retrieval BART model was judged to have the least relevant external information but the best understandability, according to raters. Exploring the tradeoff between the amount of external information and understandability, and jointly optimizing them presents a challenging direction for future work. These results confirm the effectiveness of our dataset for improving automated models' ability to generate LLS with relevant external information added. For GPT-4, when prompted with summarization and explanation, yields the highest scores in understandability, meaning preservation, and information correctness. However, it falls short in incorporating relevant external information. This suggests that GPT-4's explanatory outputs should be meticulously vetted to prevent potential misalignment with the topic. In contrast, the embedding-based Wiki approach (BART-RAG) excels in meaning preservation, maintaining the accuracy of key information, and integrating relevant external information. However, its outputs can be challenging to comprehend. This raises the potential of synergizing the embedding-based method with LLMs to achieve the ideal lay language summary.    

\subsection{Selected examples}
\label{sec:examples}
To define the scope of background explanation, we identified three types of explanation in the dataset, as shown in Table \ref{table:tmp}. We did not aim to enumerate all possible categories. Rather, our goal was to provide some initial insights into explanation phenomena. 

The most common explanation type we observed is a \textit{definition}, including ``common'' medical words, technical terms, and abbreviations, to avoid misunderstanding. \textit{Motivation}, including prevalence, risk factors, history, etc., sustains readers' interest and establishes whether the topic under discussion meets their information needs. Providing a \textit{concrete example} allows readers to link an otherwise obscure concept with a more familiar one. For example, connecting the increasing temperature in the ocean with coral reefs makes it easy for the reader to understand the importance of the study. 

We present background explanations for the two best BART models and two GPT-4 models in Table \ref{table:tmp}. This provides evidence that the retrieval-augmented model can generate both term definitions and motivations for the main topic. However, the generated external content may not be aligned with the target (e.g. Marburg virus does not cause Ebola virus disease), highlighting the importance of improving the relevance and correctness of generated abstractive summaries as an area for future research. In addition, the models appear unable to produce illustrative examples. This ability goes beyond retrieving evidence, and appears difficult to learn.

\begin{table*}
\begin{adjustbox}{center}
\includegraphics[trim={20cm 0cm 18cm 0cm},clip,scale=0.45]{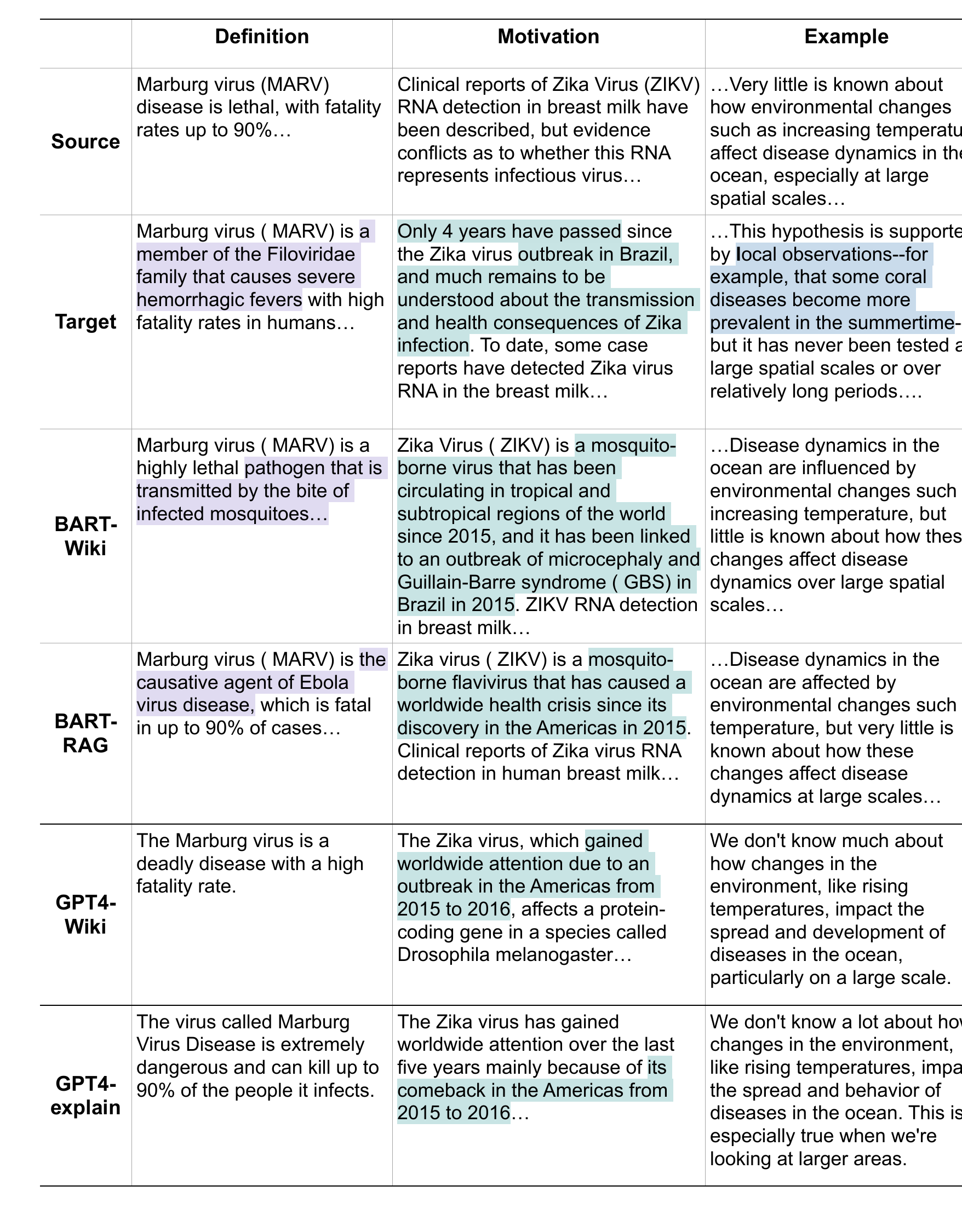}
\end{adjustbox}
 \caption{Typical types of background explanation from scientific abstracts to lay language summaries, and the corresponding generated text using two best-performing models. Our retrieval-augmented models can generate both
 term definitions and related epidemiological data
 for the main topic, but fail to link
 examples to related concepts.}
  \label{table:tmp}
\end{table*}

\section{Discussion}
We have two key observations from the model development and evaluation. Regarding BART models, RALL variants outperform the vanilla model, though these improvements are larger with the sentence simplification subset than when background explanation is emphasized. Background explanation generation is challenging and requires considering both the knowledge sources from which information is drawn, and the understandability of generated text. Examples suggest that the models can add term definitions and relevant epidemiological data but fail to provide illustrative examples for related concepts. These abilities may be beyond current models' capabilities and fall outside the scope of the resources used in our study for information retrieval. Therefore, generating high-quality explanations may require acquiring other knowledge resources, or decomposing the background explanation task into more granular subtasks. Another key observation is that human ratings are essential to assessment of background explanation task performance. Although we included automated evaluation metrics for generation quality and text simplicity, they cannot capture explanation quality. Rater evaluations of the external content for existence, relevance, and correctness of background explanations suggest additional advantages for RALL models that are opaque to automated evaluation methods.

To the best of our knowledge, CELLS is the largest lay language generation dataset developed to date, and the derived dataset for background explanation serves as the first explanation generation benchmark. We envision these data broadly applying to biomedically-related applications, and other NLP methods. On the biomedical applications side, we provide a benchmark to develop new NLP tools to generate LLSs for scientific literature. With the assistance of such NLP tools, researchers can write more understandable text, allowing healthcare consumers to interpret and apply the information it contains to guide their health-related decision-making. From an NLP methodological perspective, these datasets offer an excellent opportunity to develop and evaluate novel LLS generation techniques. CELLS can also support sentence-level and paragraph-level simplification research, and with additional annotation could provide a basis for open-question answering and informational retrieval tasks.  

While we evaluated the correctness of key information in human evaluation, it remains difficult for non-experts to identify the factuality or external information relevance. An improved model with factuality enforcement could promote sequences with higher accuracy. Medical experts (i.e. medical students) could be recruited for evaluation of factual correctness. More abstracts and human raters are required to confirm the apparent appeal of LLS from retrieval-augmented text generation models. Furthermore, to improve the quality of the dataset for background explanation, larger-scale verification is needed. We also note that our strategies for adding entity-driven explanations are straightforward, and that we did not perform a hyperparameter search to optimize the relatively expensive dense retrieval procedure, on account of resource constraints.

Evaluating lay language generation inherently poses challenges due to the multifaceted nature of the task, including aspects such as incorporating background explanations and omitting technical terms. While the ROUGE \citep{lin2004rouge} and BLEU \citep{papineni2002bleu} metrics are conventionally applied to evaluate lay language generation, their applicability is constrained due to inherent limitations associated with their reliance on lexical overlap, and the need for high-quality reference summaries. Moreover, these metrics are not adept at detecting hallucinations, a critical consideration, especially in the healthcare domain where the accuracy of lay language plays a pivotal role in informing health decisions \citep{wallace2021generating, pagnoni2021understanding}. A recent investigation indicated that ROUGE, BLEU, METEOR, and BERTScore face challenges in capturing text simplification precisely \cite{guo2023appls}. Additionally, Mac et al. found that automated readability scores frequently display inconsistency and lack accuracy \cite{mac2022comparison}. While human evaluations provide comprehensive feedback, they are resource-intensive, making them challenging to scale to extensive datasets. Therefore, a metric tailored specifically for lay language generation is much needed, and one should exercise caution when interpreting results using existing measures.

The GPSS algorithm searches for external content by calculating the lexical similarity between the source and target sentences using the ROUGE score. However, this may fail to recognize alignment at the semantic level when meanings but not terms overlap. To address this challenge, end-to-end approaches that can learn embedding-derived similarities from the source and target and classify the external content accordingly may be worth exploring. Since language models pre-trained in the medical domain have achieved state-of-the-art performance on several biomedical NLP tasks, exploring the benefits of these models is an important direction for future work. Regarding lay language generation, one remaining challenge that is a possible direction for future work involves directly applying retrieval-augmented methods to full-length abstracts instead of the background sections. Also, it would be intriguing yet challenging to generate LLS for different education levels. One potential solution may be incorporating a reward function that responds to readability, interpretability, or plainness metrics.

LLMs show promise in the realm of lay language generation. While the outputs from LLMs may not align closely with the target, the produced text is notably easy to comprehend. This ability to simplify addresses a key challenge in the existing lay language generation datasets, which typically offer only a single target. This suggests the potential to develop a pipeline that first broadens the source and then tailors the content for varying levels of readability. Moreover, the less-than-optimal performance of LLMs underscores the potential of exploring few-shot learning further. Finally, incorporating Wiki-definitions could be problematic for LLMs operating in a zero-shot learning mode. For those wishing to employ retrieval-augmented methods with LLMs, a more judicious selection of external resources or a thorough vetting of the incorporated resources is imperative.

\section{Conclusion}
To improve the interpretability of lay language text generated by neural language models, we applied state-of-the-art text generation models augmented with retrieval components and achieved promising quality and readability scores as compared with reference lay language summaries generated by human experts. Results from human evaluation support the benefits of retrieval-augmented lay language generation for the generation of background explanations in particular. The new dataset and results provide a foundation for advancement in the challenging area of automated background explanation generation and lay language generation, with the potential to mediate clearer communication of biomedical science for better informed health-related decision making.

\section{Acknowledgments}
This research was supported by US National Library of Medicine [grant number R21LM013934].



\clearpage
\newpage
\bibliographystyle{elsarticle-num-names}
\bibliography{main}





\clearpage
\newpage
\appendix
\section{Appendix}
\renewcommand{\figurename}{Appendix Figure}
\renewcommand{\tablename}{Appendix Table}
\setcounter{figure}{0}
\setcounter{table}{0}


\subsection{Background explanation annotation}
\label{background_annotation}
We provided annotators with the original abstract/LLS pair and the content (both matched and unmatched) before the 1st, 2nd, and 3rd matched sentence pairs. To make sure we have matching content from the corresponding LLS, the 1st matched sentence was included to capture situations in which explanation is provided before the first matching sentences (e.g. an introductory sentence defining terms). Examples of matching strategies can be found in Figure \ref{fig:GPSS}. We asked annotators to identify whether the filtered content 1) is in the background section (to confirm our heuristics indeed identify these sections); 2) truly contains external content (to confirm that unaligned GPSS sentences represent content that is absent from the source document); and 3) is paired (to confirm aligned GPSS sentences represent the same content). The results are shown in Appendix Table \ref{tab:background_annotation}.

\subsection{Human evaluation questions}
\label{human_evaluation_question}
The questions we included in the human evaluation questionnaire are as follows:
\begin{itemize}
    \item Is the grammar of the plain text correct? 
    \item Is the plain text easier to understand than the original text? 
    \item Does the plain text provide all the important information from the original text?
    \item Is the information in the plain text correct compared to the original text?
    \item Does the plain text provide relevant additional information compared to the original text?
\end{itemize}

\newpage
\begin{table}[]
\centering
\begin{tabular}{lcc}
\toprule
           & \textbf{Annotator 1} & \textbf{Annotator 2} \\
\midrule
\textbf{1st Pair}   &             &             \\
Background & 48          & 48          \\
External   & 20          & 18          \\
Pair       & 43          & 40          \\
\textbf{2nd Pair}   &             &             \\
Background & 47          & 47          \\
External   & 36          & 28          \\
Pair       & 47          & 46          \\
\textbf{3rd Pair}   &             &             \\
Background & 21          & 23          \\
External   & 35          & 28          \\
Pair       & 50          & 47     \\
\bottomrule
\end{tabular}
\caption{Background extraction annotation results. The columns show the number of the 50 annotated examples that each annotator labeled as containing content from the background section, including content external to the source, and being aligned with content from the source.  }
\label{tab:background_annotation}
\end{table}



\begin{table*}[t!]
\centering
\footnotesize
\begin{adjustbox}{center}
\begin{tabular}{@{}m{4.5cm}<{\centering}m{1cm}<{\centering}m{3cm}<{\centering}m{3cm}<{\centering}m{2.3cm}<{\centering} m{1.2cm}<{\centering}@{}}
\toprule
\textbf{Publisher}  & \textbf{Type}   & \textbf{Name}  & \textbf{Where are they displayed} & \textbf{Written by} & \textbf{Start from}\\ \midrule
Annals of the Rheumatic Diseases & Journal & Patient summary & Dedicated section of website & Authors & 2013
\\ \midrule
Autism & Journal & Lay abstract & Dedicated section of website & Authors & 2011
\\ \midrule
Autism Research & Journal & Lay abstract & Dedicated section of website & Authors & 2008
 \\ \midrule
British Journal of Dermatology & Journal & QAs & Issues searching page & Authors/editorial team & 2013
\\ \midrule
Cochrane & Journal & Plain language summary & Within article & Authors/editorial team & 1997
\\ \midrule
eLife & Blog & eLife digest & Section on blog & Editorial team & 2012
\\ \midrule
European Urology & Journal & Patient summary & Within article & Authors & 2014
\\ \midrule
NIHR Efficaacy and Mechanism Evaluation & Journal & Plain English summary & Within article & Authors & 2014
\\ \midrule
NIHR Health Services and Delivery Response & Journal & Plain English summary & Within article & Authors & 2014
\\ \midrule
NIHR Health Technology Assessment & Journal & Plain English summary & Within article & Authors & 2014
\\ \midrule
NIHR Programme Grants for Applied Research & Journal & Plain English summary & Within article & Authors & 2015
\\ \midrule
PLOS Biology & Journal & Author summary & Within article & Authors & 2007
\\ \midrule
PLOS Computational Biology & Journal & Author summary & Within article & Authors & 2005
\\ \midrule
PLOS Genetics & Journal & Author summary & Within article & Authors & 2005
\\ \midrule
PLOS Medicine & Journal & Author summary & Within article & Authors & 2006
\\ \midrule
PLOS Neglected Tropical Diseases & Journal & Author summary & Within article & Authors & 2007
\\ \midrule
PLOS Pathogens & Journal & Author summary & Within article & Authors & 2005
\\ \midrule
Proceedings of the National Academy of Sciences & Journal & Significance & Within article & Authors & 2013
\\ \midrule
Reproductive Health & Journal & Plain English summary & Within article & Authors & 2016
\\\bottomrule
\end{tabular}
\end{adjustbox}
\caption{Summary of journals with Plain Language Summary}
\label{journal_pls}
\end{table*}

\begin{figure*}[h!]
    \centering
    \begin{adjustbox}{center}
    \includegraphics[]{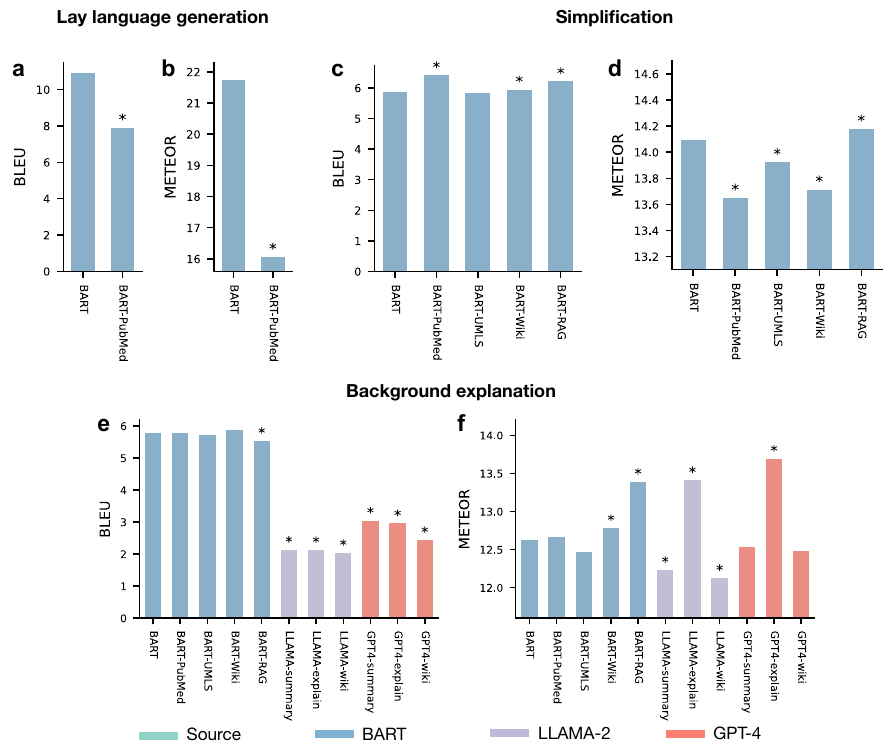}
    \end{adjustbox}
    \caption{Models' performance in text generation. We used the F1 score of BLEU and METEOR to evaluate the generation quality of models on lay language generation, simplification, and background explanation tasks. P-values obtained through the t-test are employed to evaluate the performance of various models compared to the Vanilla model (BART). A p-value less than 0.05 is indicated by (*).}
    \label{appfig:generation_quality_bleu}
    \vspace{-.25cm}
\end{figure*}

\begin{figure*}[h!]
    \centering
    \begin{adjustbox}{center}
    \includegraphics[]{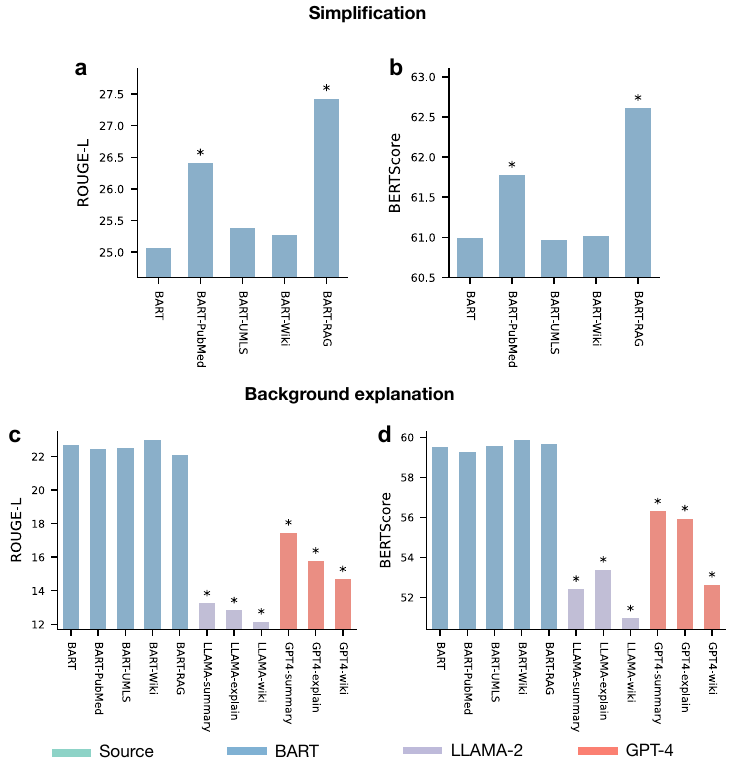}
    \end{adjustbox}
    \caption{Models' performance in text generation on the validated dataset. We used the F1 score of ROUGE-L and BERTScore to evaluate the generation quality of models on lay language generation, simplification, and background explanation tasks. P-values obtained through the t-test are employed to evaluate the performance of various models compared to the Vanilla model (BART). A p-value less than 0.05 is indicated by (*).}
    \label{appfig:generation_quality_validate_rouge}
    \vspace{-.25cm}
\end{figure*}

\begin{figure*}[h!]
    \centering
    \begin{adjustbox}{center}
    \includegraphics[]{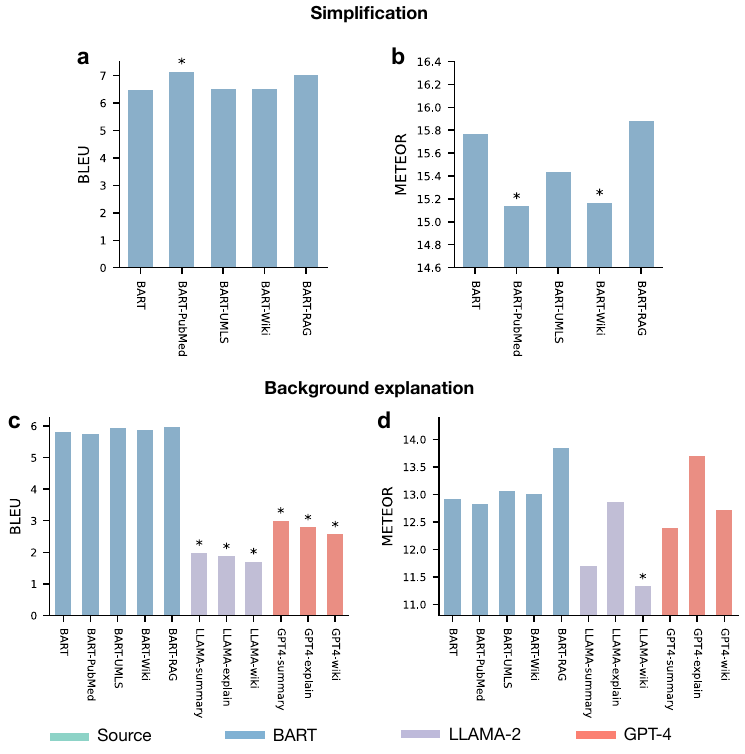}
    \end{adjustbox}
    \caption{Models' performance in text generation on the validated dataset. We used the F1 score of BLEU and METEOR to evaluate the generation quality of models on lay language generation, simplification, and background explanation tasks. P-values obtained through the t-test are employed to evaluate the performance of various models compared to the Vanilla model (BART). A p-value less than 0.05 is indicated by (*).}
    \label{appfig:generation_quality_validate_bleu}
    \vspace{-.25cm}
\end{figure*}

\begin{figure}[h!]
    \centering
    \includegraphics[scale=1]{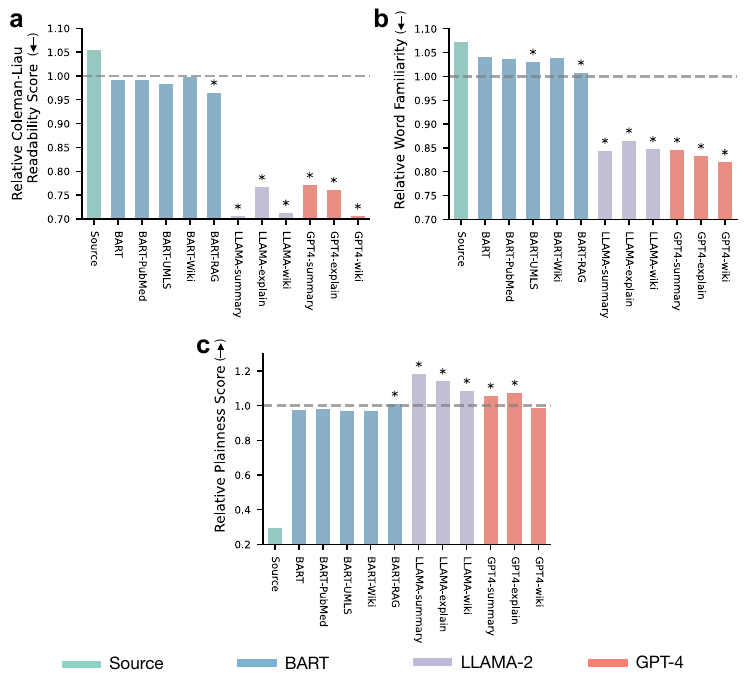}
    \caption{Readibility, familiarity, and plainness of the validated background explanation subset. (a) Relative Coleman-Liau readability score, (b) word familiarity, and (c) plainness score of the source and models' generated text. The relative score is calculated by dividing by the score of the target text. A lower readability score and word familiarity indicate that the text is easier to read (values below the dashed line are lower than those from professionally-authored plain language). A higher Plainness Score indicates that the text is more representative of an LLS. P-values obtained through the t-test are employed to evaluate the performance of various models compared to the Vanilla model (BART). A p-value less than 0.05 is indicated by (*).}
    \label{appfig:test_results_validated}
    \vspace{-.25cm}
\end{figure}
\end{document}